\pdfoutput=1
\documentclass[11pt,a4paper]{article}
\usepackage[hyperref]{acl2019}
\usepackage{times}
\usepackage{latexsym}
\usepackage{tabularx}

\usepackage{tikz}
\definecolor{gitred}{HTML}{FDB8C0}
\definecolor{gitgreen}{HTML}{006400}
\definecolor{chocolate}{HTML}{D2691E}
\definecolor{maroon}{HTML}{800000}
\definecolor{indigo}{HTML}{4B0082}
\definecolor{green}{HTML}{008000}
\usepackage{pifont}

\usepackage[normalem]{ulem}

\usepackage{enumitem}
\usepackage{booktabs}
\usepackage{multirow}
\usepackage{graphicx}
\usepackage{array}
\usepackage{subfigure}
\usepackage{amsmath}
\usepackage{url}
\usepackage{makecell}
\usepackage{amsmath}
\usepackage{amsfonts}
\usepackage{amsthm,amssymb,amsopn,bm}
\usepackage{color,soul}
\usepackage{verbatim}
\usepackage{algorithm}
\usepackage[noend]{algpseudocode}
\DeclareMathOperator*{\argmax}{argmax}

\usepackage{tabulary}
\usepackage{footnote}
\makesavenoteenv{algorithm}

\DeclareRobustCommand{\hlcyan}[1]{{\sethlcolor{cyan}\hl{#1}}}
\DeclareRobustCommand{\hlorange}[1]{{\sethlcolor{orange}\hl{#1}}}
\DeclareRobustCommand{\hlpink}[1]{{\sethlcolor{pink}\hl{#1}}}

\aclfinalcopy 
\def\aclpaperid{1917} 

\providecommand{\sewon}[1]{
    {\protect\color{blue}{[Sewon: #1]}}
}
\providecommand{\victor}[1]{
    {\protect\color{purple}{[Victor: #1]}}
}
\providecommand{\hanna}[1]{
    {\protect\color{indigo}{\bf [Hanna: #1]}}
}

\title{Multi-hop Reading Comprehension \\ through Question Decomposition and Rescoring}

\author{
    Sewon Min$^1$, Victor Zhong$^1$, Luke Zettlemoyer$^{1}$, Hannaneh Hajishirzi$^{1,2}$ \\
    $^1$University of Washington\\
    $^2$Allen Institute for Artificial Intelligence\\
    {\tt $\{$sewon,vzhong,lsz,hannaneh$\}$@cs.washington.edu}
}

\date{}
\begin{document}

\newcommand{\bert}{\textsc{BERT}}
\newcommand{\bertbase}{\textsc{BERT-BASE}}
\providecommand{\pointer}[1]{$\mathrm{Pointer_#1}$}

\newcommand{\hotpot}{\textsc{HotpotQA}}
\newcommand{\wikihop}{\textsc{WikiHop}}
\newcommand{\cwq}{\textsc{ComplexWebQuestions}}
\newcommand{\wq}{\textsc{WebQuestions}}
\newcommand{\squad}{\textsc{SQuAD}}
\newcommand{\sys}{\textsc{DecompRC}}

\newcommand{\answer}{\texttt{ANS}}
\newcommand{\keyword}{\texttt{ENT}}
\newcommand{\whword}{\textsl{wh}-word}

\newcommand{\sota}{the state-of-the-art}
\newcommand{\rc}{reading comprehension}
\newcommand{\RClong}{Reading Comprehension}
\newcommand{\RC}{RC}
\newcommand{\query}{sub-question}
\newcommand{\queries}{sub-questions}
\newcommand{\Query}{Sub-question}
\newcommand{\Queries}{Sub-questions}

\newcommand{\yes}{\texttt{yes}}
\newcommand{\no}{\texttt{no}}
\providecommand{\example}[1]{``#1"}

\maketitle

\begin{abstract}
    Multi-hop \RClong{} (\RC) requires reasoning and aggregation across several paragraphs.
    We propose a system for multi-hop \RC{} that decomposes a compositional question into simpler \queries{} that can be answered by off-the-shelf single-hop \RC{} models.
    Since annotations for such decomposition are expensive, we recast sub-question generation as a span prediction problem and 
    show that our method, trained using only 400 labeled examples, generates \queries{} that are as effective as human-authored \queries{}.
    We also introduce a new global rescoring approach that considers each decomposition (i.e. the \queries{} and their answers) to select the best final answer, greatly improving overall performance.
    Our experiments on \hotpot{} show that this approach achieves \sota{} results, while providing explainable evidence for its decision making in the form of \queries{}.
\end{abstract}
\section{Introduction}\label{sec:intro}
Multi-hop \rc{} (\RC) is challenging because it requires the aggregation of evidence across several paragraphs to answer a question.
Table~\ref{tab:intro-example} shows an example of multi-hop \RC{}, where the question \example{Which team does the player named 2015 Diamond Head Classics MVP play for?} requires first finding the player who won MVP from one paragraph, and then finding the team that player plays for from another paragraph.

\begin{table}[t]
\begin{center}
\resizebox{\columnwidth}{!}{
\begin{tabular}{@{}l@{}} 
 \toprule
    {\textbf{Q}} Which team does \hl{the player named 2015 Diamond Head} \\
    \hl{Classic's MVP} play for? \\
    {\textbf{P1}} The 2015 Diamond Head Classic was ... Buddy Hield was\\
    named the tournament's MVP. \\
    {\textbf{P2}} Chavano Rainier Buddy Hield is a Bahamian professional \\
    basketball player for the {\protect\color{red}{Sacramento Kings}} ... \\
    \midrule
    {\textbf{Q1}} Which player named 2015 Diamond Head Classic's MVP? \\
    {\textbf{Q2}} Which team does \answer~play for? \\
 \bottomrule
\end{tabular}
}
\end{center}
\caption{An example of multi-hop question from \hotpot{}. The first cell shows given question and two of given paragraphs (other eight paragraphs are not shown), where the {\protect\color{red}{red text}} is the groundtruth answer. Our system selects a \hl{span} over the question and writes two \queries{} shown in the second cell.}
\label{tab:intro-example}
\vspace{-10pt}
\end{table}

In this paper, we propose \sys, a system for multi-hop \RC, that learns to break compositional multi-hop questions into simpler, single-hop \queries{} using spans from the original question.  For example, for the question in Table~\ref{tab:intro-example}, we can create the \queries{}~\example{Which player named 2015 Diamond Head Classics MVP?}~and~\example{Which team does \answer{} play for?}, where the token \answer{} is replaced by the answer to the first \query{}. 
The final answer is then the answer to the second \query{}.



Recent work on question decomposition relies on distant supervision data created on top of underlying relational logical forms~\citep{cwq}, making it difficult to generalize to diverse natural language questions such as those on \hotpot~\citep{hotpot}.
In contrast, our method presents a new approach which simplifies the process as a span prediction, thus requiring only 400 decomposition examples to train a competitive decomposition neural model.
Furthermore, we propose a rescoring approach which obtains answers from different possible decompositions and rescores each decomposition with the answer to decide on the final answer, rather than deciding on the decomposition in the beginning.



Our experiments show that \sys\ outperforms other published methods on \hotpot~\citep{hotpot}, while providing explainable evidence in the form of \queries.
In addition, we evaluate with alternative distrator paragraphs and questions and show that our decomposition-based approach is more robust than an end-to-end \bert{} baseline~\citep{bert}.
Finally, our ablation studies show that our \queries{}, with 400 supervised examples of decompositions, are as effective as human-written \queries, and that our answer-aware rescoring method significantly improves the performance.

Our code and interactive demo are publicly available at \url{https://github.com/shmsw25/DecompRC}.
\begin{figure*}[!htb]
\centering
\includegraphics[width=\textwidth]{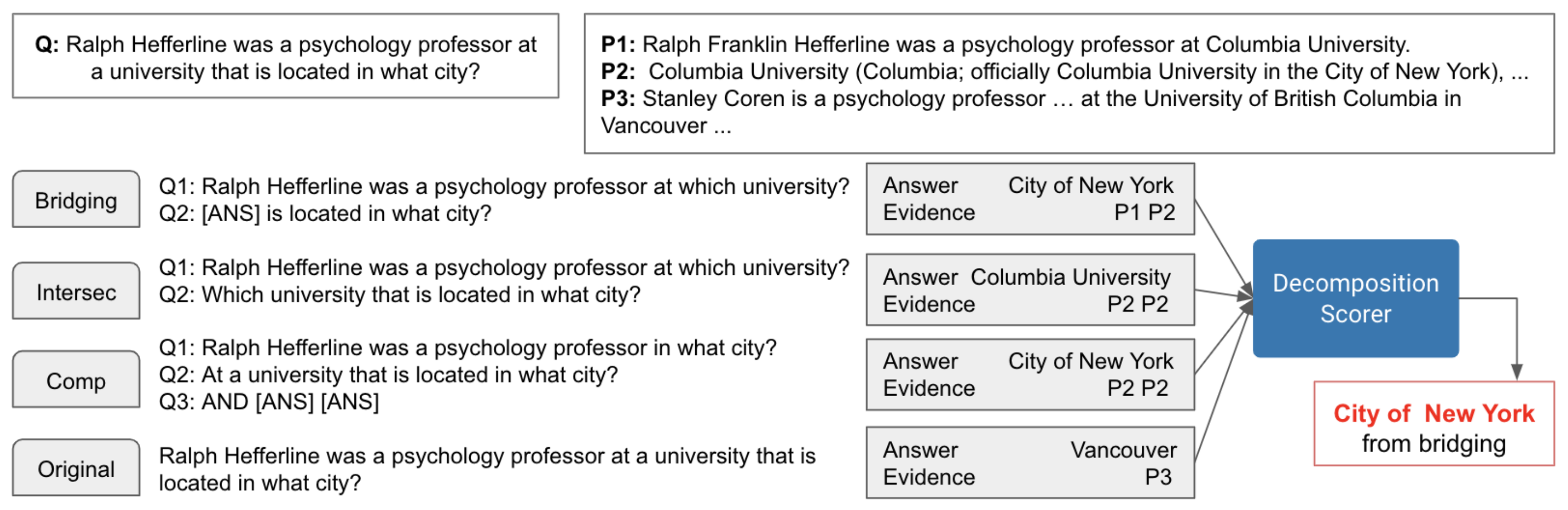}
\caption{
The overall diagram of how our system works. Given the question, \sys\ decomposes the question via all possible reasoning types (Section~\ref{subsec:decomposition}). Then, each \query{} interacts with the off-the-shelf \RC{} model and produces the answer (Section~\ref{subsec:single}). Lastly, the decomposition scorer decides which answer will be the final answer (Section~\ref{subsec:recombination}). Here, \example{City of New York}, obtained by bridging, is determined as a final answer.
}
\label{fig:briging-example}
\end{figure*}


\section{Related Work}\label{sec:related}
\paragraph{\RClong.}
In reading comprehension, a system reads a document and answers questions regarding the content of the document~\citep{mctest}.
Recently, the availability of large-scale \rc datasets~\citep{cnndailymail, squad, triviaqa} has led to the development of advanced \RC{} models~\citep{bidaf, dcn+, fast-and-accurate, bert}.
Most of the questions on these datasets can be answered in a single sentence~\citep{minimal}, which is a key difference from multi-hop \rc.

\paragraph{Multi-hop \RClong.}
In multi-hop \rc, the evidence for answering the question is scattered across multiple paragraphs.
Some multi-hop datasets contain questions that are, or are based on relational queries~\citep{wikihop,cwq}.
In contrast, \hotpot{}~\citep{hotpot}, on which we evaluate our method, contains more natural, hand-written questions that are not based on relational queries.

Prior methods on multi-hop \rc{} focus on answering relational queries, and emphasize attention models that reason over coreference chains~\citep{dhingra2018neural,zhong2019wikihop,cao2018wikihop}.
In contrast, our method focuses on answering natural language questions via question decomposition.
By providing decomposed single-hop \queries{}, our method allows the model's decisions to be explainable.

Our work is most related to~\citet{cwq}, which answers questions over web snippets via decomposition.
There are three key differences between our method and theirs.
First, they decompose questions that are correspond to relational queries, whereas we focus on natural language questions.
Next, they rely on an underlying relational query (SPARQL) to build distant supervision data for training their model, while our method requires only 400 decomposition examples.
Finally, they decide on a decomposition operation exclusively based on the question.
In contrast, we decompose the question in multiple ways, obtain answers, and determine the best decomposition based on all given context, which we show is crucial to improving performance.

\paragraph{Semantic Parsing.}
Semantic parsing is a larger area of work that involves producing logical forms from natural language utterances, which are then usually executed over structured knowledge graphs~
\citep{zelle96LearnToParse,Zettlemoyer2005LearningTM,Liang2011LearningDependencyBasedCompositionalSemantics}.
Our work is inspired by the idea of compositionality from semantic parsing, however, we focus on answering natural language questions over unstructured text documents.
\section{Model}
\subsection{Overview}~\label{subsec:overview}
In multi-hop \rc{}, a system answers a question over a collection of paragraphs by combining evidence from multiple paragraphs.
In contrast to single-hop \rc{}, in which a system can obtain good performance using a single sentence~\citep{minimal}, multi-hop \rc{} typically requires more complex reasoning over how two pieces of evidence relate to each other. 

We propose \sys{} for multi-hop \rc{} via question decomposition.
\sys\ answers questions through a three step process:

\begin{table*}[t]
    \centering
    \small
    \begin{tabularx}{\textwidth}{@{}cX@{}}
        \toprule     
         \textbf{Type} & \textbf{Bridging (47\%)} requires finding the first-hop evidence in order to find another, second-hop evidence. \\ 
         {Q} & Which team does \hl{the \textbf{player} named 2015 Diamond Head Classic's MVP} play for? \\
         {Q1} & Which player named 2015 Diamond Head Classic's MVP? \\
         {Q2} & Which team does \answer~play for? \\
        \midrule
         \textbf{Type} & \textbf{Intersection (23\%)} requires finding an entity that satisfies two independent conditions. \\ 
         {Q} & Stories USA starred {\protect\color{red}{\checkmark}} which actor and comedian {\protect\color{red}{\checkmark}} from `The Office'?\\
         {Q1} & Stories USA starred which actor and comedian? \\
         {Q2} & Which actor and comedian from `The Office'? \\
        \midrule
         \textbf{Type}& \textbf{Comparison (22\%)} requires comparing the property of two different entities. \\
         {Q} & Who was born earlier, \hl{Emma Bull}  or \hl{Virginia Woolf}? \\
         {Q1} & Emma Bull was born when? \\
         {Q2} & Virginia Woolf was born when? \\
         {Q3} & Which\_is\_smaller (Emma Bull,~\answer) (Virgina Woolf,~\answer) \\
        \bottomrule              
    \end{tabularx}
    \vskip -0.7em
    \caption{The example multi-hop questions from each category of reasoning type on~\hotpot.
    Q indicates the original, multi-hop question, while Q1, Q2 and Q3 indicate \queries{}.
    \sys\ predicts \hl{span} and {\protect\color{red}{\checkmark}} through \pointer{c}, generates \queries{}, and answers them iteratively through single-hop \RC{} model.
    }
    \label{tab:decomposition-example}
\end{table*}

\begin{enumerate}\itemsep0em 
  \item
    First, \sys\ decomposes the original, multi-hop {\em question} into several single-hop {\em \queries{}} according to a few {\em reasoning types}  in parallel, based on {\em span} predictions.
    Figure~\ref{fig:briging-example} illustrates an example in which a question is decomposed through four different reasoning types.
    Section~\ref{subsec:decomposition} details our decomposition approach.
  \item
    Then, for every reasoning types \sys\ leverages a single-hop \rc{} model to answer each \query{}, and combines the answers according to the reasoning type.
    Figure~\ref{fig:briging-example} shows an example for which bridging produces `City of New York' as an answer while intersection produces `Columbia University' as an answer. 
    Section~\ref{subsec:single} details the single-hop \rc{} procedure.
  \item
    Finally, \sys\ leverages a {\em decomposition scorer} to judge which decomposition is the most suitable, and outputs the answer from that decomposition as the {\em final answer.}
    In Figure~\ref{fig:briging-example}, \example{City of New York}, obtained via bridging, is decided as the final answer.
    Section~\ref{subsec:recombination} details our rescoring step.
\end{enumerate}
We identify several reasoning types in multi-hop \rc{}, which we use to decompose the original question and rescore the decompositions.
These reasoning types are {\em bridging}, {\em intersection} and {\em comparison}.
Table~\ref{tab:decomposition-example} shows examples of each reasoning type.
On a sample of 200 questions from the dev set of \hotpot{}, we find that 92\% of multi-hop questions belong to one of these types.
Specifically, among 184 samples out of 200 which require multi-hop reasoning, 47\% are bridging questions, 23\% are intersection questions, 22\% are comparison questions, and 8\% do not belong to one of three types.
In addition, these multi-hop reasoning types correspond to the types of compositional questions identified by~\citet{webquestions} and~\citet{cwq}.

\subsection{Decomposition}~\label{subsec:decomposition} 
The goal of question decomposition is to convert a multi-hop question into simpler, single-hop \queries{}.
A key challenge of decomposition is that it is difficult to obtain annotations for how to decompose questions.
Moreover, generating the question word-by-word is known to be a difficult task that requires substantial training data and is not straight-forward to evaluate~\citep{nlgsurvey,betternlgevaluation}.

Instead, we propose a method to create \queries{} using span prediction over the question.
The key idea is that, in practice, each \query{} can be formed by copying and lightly editing a key span from the original question, with different span extraction and editing required for each reasoning type.
For instance, the bridging question in Table~\ref{tab:decomposition-example} requires finding \example{the player named 2015 Diamond Head Classic MVP} which is easily extracted as a span.
Similarly, the intersection question in Table~\ref{tab:decomposition-example} specifies the type of entity to find (\example{which actor and comedian}), with two conditions (\example{Stories USA starred} and \example{from ``The Office''}), all of which can be extracted.
Comparison questions compare two entities using a discrete operation over some properties of the entities, e.g., \example{which is smaller}. When two entities are extracted as spans, the question can be converted into two \queries{}  and one discrete operation over the answers of the \queries{}.
 
\paragraph{Span Prediction for \Query{} Generation} Our approach simplifies the \query{} generation problem into a span prediction problem that requires little supervision (400 annotations).
The annotations are collected by mapping the question into several points that segment the question into spans (details in Section~\ref{subsec:train-details}).
We train a model \pointer{c} that learns to map a question into $c$ points, which are subsequently used to compose \queries{} for each reasoning type through Algorithm~\ref{alg:qg}. 

\pointer{c} is a function that points to $c$ indices $\mathrm{ind}_1, \dots, \mathrm{ind}_c$ in an input sequence.\footnote{$c$ is a hyperparameter which differs in different reasoning types.}
Let $S = [s_1, \dots, s_n]$ denote a sequence of $n$ words in the input sequence.
The model encodes $S$ using \bert{}~\citep{bert}: \begin{equation}\label{eq:bert}
    U = \bert (S) \in \mathbb{R}^{n \times h},
\end{equation}
where $h$ is the output dimension of the encoder.
\begin{algorithm}
\begin{algorithmic}
\small
\Procedure{GenerateSubQ}{$Q:\text{question}$, \pointer{c}}
\State{{\protect\color{gitgreen}{\textit{/* Find $q^b_1$ and $q^b_2$ for Bridging */}}}}
\State $\mathrm{ind}_1, \mathrm{ind}_2, \mathrm{ind}_3  \gets$ \pointer{3}$(Q)$ 
\State $q^b_1 \gets Q_{\mathrm{ind}_1:\mathrm{ind}_3}$ 
\State $q^b_2 \gets Q_{:\mathrm{ind}_1}:\answer:Q_{\mathrm{ind}_3:}$ 
\State {article in $Q_{\mathrm{ind}_2-5:\mathrm{ind}_2} \gets$ `which'}
\State{{\protect\color{gitgreen}{\textit{/* Find $q^i_1$ and $q^i_2$ for Intersecion */}}}}
\State $\mathrm{ind}_1, \mathrm{ind}_2 \gets$ \pointer{2}$(Q)$ 
\State $s_1, s_2, s_3 \gets Q_{:\mathrm{ind}_1}, Q_{\mathrm{ind}_1:\mathrm{ind}_2}, Q_{\mathrm{ind}_2:}$ 
\If{$s_2$ starts with wh-word}
    \State $q^i_1 \gets s_1:s_2,~q^i_2 \gets s_2:s_3$
\Else
    \State $q^i_1 \gets s_1:s_2,~q^i_2 \gets s_1:s_3$
\EndIf
\State{{\protect\color{gitgreen}{\textit{/* Find $q^c_1$, $q^c_2$ and $q^c_3$ for Comparison */}}}}
\State $\mathrm{ind}_1, \mathrm{ind}_2, \mathrm{ind}_3, \mathrm{ind}_4 \gets $\pointer{4}$(Q)$
\State $\text{ent}_1, \text{ent}_2 \gets Q_{\mathrm{ind}_1:\mathrm{ind}_2}, Q_{\mathrm{ind}3:\mathrm{ind}_4}$
\State $op \gets \mathrm{find\_op}(Q, \text{ent}_1, \text{ent}_2)$ 
\State $q^c_1$, $q^c_2 \gets \mathrm{form\_subq}(Q, \text{ent}_1, \text{ent}_2, op)$  
\State $q^c_3 \gets op~(\text{ent}_1, \answer)~(\text{ent}_2, \answer)$
\EndProcedure
\end{algorithmic}
\caption{\Queries{} generation using \pointer{c}.\footnote{Details for $ \mathrm{find\_op}$, $\mathrm{form\_subq}$ in Appendix~\ref{app:heuristics}.}
}\label{alg:qg}
\end{algorithm}

Let $W \in \mathbb{R}^{h \times c}$ denote a trainable parameter matrix.
We compute a pointer score matrix \begin{equation}
    Y = \mathrm{softmax}(U W) \in \mathbb{R}^{n \times c},
\end{equation} where $\mathbb{P}(i = \mathrm{ind}_j) = Y_{ij}$ denotes the probability that the $i$th word is the $j$th index produced by the pointer.
The model extracts $c$ indices that yield the highest joint probability at inference: \begin{align*}
    \mathrm{ind}_1, \dots, \mathrm{ind}_c  = \argmax_{i_1 \leq \dots \leq {i}_c} \prod_{j=1}^{c} \mathbb{P}(i_j=\mathrm{ind}_j)
\end{align*}

\subsection{Single-hop \RClong}~\label{subsec:single}
Given a decomposition, we use a single-hop \RC{} model to answer each \query{}.
Specifically, the goal is to obtain the answer and the evidence, given the \query{} and $N$ paragraphs.
Here, the answer is a span from one of paragraphs, \texttt{yes} or \texttt{no}.
The evidence is one of $N$ paragraphs on which the answer is based.

Any off-the-shelf \RC{} model can be used.
In this work, we use the \bert{} \rc{} model~\citep{bert} combined with the paragraph selection approach from~\citet{clark2018multi} to handle multiple paragraphs.
Given $N$ paragraphs $S_1, \dots, S_N$, this approach independently computes $\mathrm{answer}_i$ and $y_i^\mathrm{none}$ from each paragraph $S_i$, where $\mathrm{answer}_i$ and $y_i^\mathrm{none}$ denote the answer candidate from $i$th paragraph and the score indicating $i$th paragraph does not contain the answer.
The final answer is selected from the paragraph with the lowest $y_i^\mathrm{none}$. Although this approach takes a set of multiple paragraphs as an input, it is not capable of jointly reasoning across different paragraphs.

For each paragraph $S_i$, let $U_{i} \in \mathbb{R}^{n \times h}$ be the \bert{} encoding of the \query{} concatenated with a paragraph $S_i$, obtained by Equation~\ref{eq:bert}.
We compute four scores, $y_i^\mathrm{span}$ $y_i^\mathrm{yes}$, $ y_i^\mathrm{no}$ and $y_i^\mathrm{none}$, indicating if the answer is a phrase in the paragraph, \texttt{yes}, \texttt{no}, or does not exist.
\begin{align*} [y_i^\mathrm{span}; y_i^\mathrm{yes}; y_i^\mathrm{no}; y_i^\mathrm{none}]  = \mathrm{max}(U_i)  W_1 \in \mathbb{R}^4, \end{align*}
where $\max$ denotes a max-pooling operation across the input sequence, and $W_1 \in \mathbb{R}^{h \times 4}$ denotes a parameter matrix.
Additionally, the model computes $ \mathrm{span}_i$, which is defined by its start and end points $\mathrm{start}_i$ and $\mathrm{end}_i$.
\begin{eqnarray*}
    \mathrm{start}_i, \mathrm{end}_i &=& \argmax_{j \leq k} \mathbb{P}_{i, \mathrm{start}}(j)\mathbb{P}_{i, \mathrm{end}}(k), 
\end{eqnarray*}
where $\mathbb{P}_{i, \mathrm{start}}(j)$ and $\mathbb{P}_{i, \mathrm{end}}(k)$ indicate the probability that the $j$th word is the start and the $k$th word is the end of the answer span, respectively.
$\mathbb{P}_{i, \mathrm{start}}(j)$ and $\mathbb{P}_{i, \mathrm{end}}(k)$ are obtained by the $j$th element of ${p}^\mathrm{start}_i$ and the $k$th element of ${p}^\mathrm{end}_i$ from
\begin{eqnarray}
    {p}^\mathrm{start}_i &=& \mathrm{softmax}(U_i  W_\mathrm{start}) \in \mathbb{R}^{n} \\
    {p}^\mathrm{end}_i &=& \mathrm{softmax}(U_i  W_\mathrm{end}) \in \mathbb{R}^{n}
\end{eqnarray}
Here, $W_\mathrm{start}, W_\mathrm{end} \in \mathbb{R}^h$ are the parameter matrices.
Finally, $\mathrm{answer}_i$ is determined as one of $\mathrm{span}_i$, \texttt{yes} or \texttt{no} based on which of $y_i^\mathrm{span}$, $y_i^\mathrm{yes}$ and $ y_i^\mathrm{no}$ is the highest. 

The model is trained using questions that only require single-hop reasoning, obtained from \squad{}~\citep{squad} and easy examples of \hotpot~\citep{hotpot} (details in Section~\ref{subsec:train-details}). Once trained, it is used as an off-the-shelf RC model and is never directly trained on multi-hop questions. 

\subsection{Decomposition Scorer}\label{subsec:recombination}
Each decomposition consists of \queries{}, their answers, and evidence corresponding to a reasoning type.
\sys\ scores decompositions and takes the answer of the top-scoring decomposition to be the final answer.  
The score indicates if a decomposition leads to a correct final answer to the multi-hop question.

Let $t$ be the reasoning type, and let $\text{answer}_t$ and $\text{evidence}_t$ be the answer and the evidence from the reasoning type $t$.
Let $x$ denote a sequence of $n$ words formed by the concatenation of the question, the reasoning type $t$, the answer $\text{answer}_t$, and the evidence $\text{evidence}_t$.
The decomposition scorer encodes this input $x$ using BERT to obtain $U_{t} \in \mathbb{R}^{n \times h}$ similar to Equation~\eqref{eq:bert}.
The score $p_t$ is computed as \begin{align*} p_t = \mathrm{sigmoid} (W_2^T \mathrm{max}(U_{t})) \in \mathbb{R}, \end{align*} where $W_2 \in \mathbb{R}^{h}$ is a trainable matrix. 

During inference, the reasoning type is decided as $\argmax_t p_t$.
The answer corresponding to this reasoning type is chosen as the final answer.


\paragraph{Pipeline Approach.}
An alternative to the decomposition scorer is a pipeline approach, in which the reasoning type is determined in the beginning, before decomposing the question and obtaining the answers to sub-questions.
Section~\ref{subsec:ablation} compares our scoring step with this approach to show the effectiveness of the decomposition scorer. Here, we briefly describe the model used for the pipeline approach.

First, we form a sequence $S$ of $n$ words from the question and obtain $\tilde{S} \in \mathbb{R}^{n \times h}$ from Equation~\ref{eq:bert}. Then, we compute 4-dimensional vector $p_t$ by: \begin{align*} p_t = \mathrm{softmax} (W_3 \mathrm{max}(\tilde{S})) \in \mathbb{R}^4 \end{align*} where $W_3 \in \mathbb{R}^{h \times 4}$ is a parameter matrix. Each element of 4-dimensional vector $p_t$ indicates the reasoning type is bridging, intersection, comparison or original.

\section{Experiments}\label{sec:exp}

\subsection{\hotpot}

We experiment on \hotpot{}~\citep{hotpot}, a recently introduced multi-hop \RC{} dataset over Wikipedia articles. There are two types of questions---bridge and comparison. Note that their categorization is based on the data collection and is different from our categorization (bridging, intersection and comparison) which is based on the required reasoning type. 
We evaluate our model on dev and test sets in two different settings, following prior work.

\vspace{.1cm}
\noindent \textbf{Distractor setting} contains the question and a collection of 10 paragraphs: 2 paragraphs are provided to crowd workers to write a multi-hop question, and 8 distractor paragraphs are collected separately via TF-IDF between the question and the paragraph.
The train set contains easy, medium and hard examples, where easy examples are single-hop, and medium and hard examples are multi-hop. The dev and test sets are made up of only hard examples.

\vspace{.1cm}
\noindent \textbf{Full wiki setting} is an open-domain setting which contains the same questions as distractor setting but does not provide the collection of paragraphs.
Following \citet{squad-open}, we retrieve 30 Wikipedia paragraphs based on TF-IDF similarity between the paragraph and the question (or \query).

\begin{table*}[tb]
    \centering
    \begin{tabulary}{\textwidth}{p{2cm}|ccccc|ccccc} 
     \toprule
        & \multicolumn{5}{c|}{\em \textbf{Distractor setting}} & \multicolumn{5}{c}{\em \textbf{Full wiki setting}} \\
        & {All} & {Bridge} & {Comp} & Single & Multi  & {All} & {Bridge} & {Comp} & Single & Multi \\
     \midrule
        {\sys} & \textbf{70.57} & \textbf{72.53} & \textbf{62.78} & 84.31 & \textbf{58.74}  & \textbf{43.26} & \textbf{40.30} & \textbf{55.04} & \textbf{52.11} & \textbf{35.64} \\
        \ \ \ \ \ \ 1hop train &61.73 & 61.57 & 62.36 & 79.38 & 46.53 & 39.17 & 35.30 & 54.57 & 50.03 & 29.83\\
        {\bert} &67.08 & 69.41 & 57.81 & 82.98 & 53.38&38.40 & 34.77 & 52.85 & 46.14 & 31.74 \\
        \ \ \ \ \ \ 1hop train & 56.27 & 62.77 & 30.40 & \textbf{87.21} & 29.64  &29.97 & 32.15 & 21.29 & 47.14 & 15.18 \\
        BiDAF& 58.28 &  59.09 &  55.05 & - & - & 34.36 & 30.42 &  50.70 & - & - \\
     \bottomrule
\end{tabulary}
\caption{F1 scores on the dev set of \hotpot{} in both distractor  (left) and full wiki settings (right). 
We compare \sys{} (our model), \bert{}, and BiDAF, and variants of the models that are only trained on single-hop QA data ({\em 1hop train}).
{\em Bridge} and {\em Comp} indicate original splits in \hotpot{}; {\em Single} and {\em Multi} refer to dev set splits  that can be solved (or not) by all of three \bert{} models trained on single-hop QA data.} 
\label{tab:hotpot-result}
\vspace{-8pt}
\end{table*}

\begin{table}[tb]
    \centering
    \begin{tabular}{l|c|c} 
     \toprule
      Model & Dist F1 & Open F1 \\
     \midrule
        \sys & 69.63 & 40.65\\
     \midrule
        Cognitive Graph & - & {48.87} \\
        BERT Plus & {69.76} & -\\
        MultiQA & - & 40.23 \\
        DFGN+BERT & 68.49 & - \\
        QFE & 68.06 & 38.06 \\
        GRN & 66.71 & 36.48 \\
        BiDAF & 59.02 & 32.89 \\
     \bottomrule
\end{tabular}
\caption{F1 score on the test set of \hotpot{} distractor and full wiki setting. All numbers from the official leaderboard. All models except BiDAF are concurrent work (not published). \sys\ achieves the best result out of models reported to both distractor and full wiki setting.} 
\label{tab:hotpot-test}
\vspace{-8pt}
\end{table}

\subsection{Implementations Details}\label{subsec:train-details}
\paragraph{Training {\textbf{Pointer}} for Decomposition.}
We obtain a set of 200 annotations for bridging to train \pointer{3}, and another set of 200 annotations for intersection to train \pointer{2}, hence 400 in total.
Each bridging question pairs with three points in the question, and each intersection question pairs with two points in the question.
For comparison, we create training data in which each question pairs with four points (the start and end of the first entity and those of the second entity) to train \pointer{4}, requiring no extra annotation.\footnote{Details in Appendix~\ref{app:heuristics}.}

\paragraph{Training Single-hop \RC{} Model.} We create {\em single-hop QA data} by combining \hotpot{} easy examples and SQuAD~\citep{squad} examples to form the training data for our single-hop \RC{} model described in Section~\ref{subsec:single}.
To convert \squad{} to a multi-paragraph setting, we retrieve $n$ other Wikipedia paragraphs based on TF-IDF similarity between the question and the paragraph, using Document Retriever from DrQA~\citep{squad-open}.
We train 3 instances with $n=0,2,4$ for an ensemble, which we use as the single-hop model.

To deal with sections/ungrammatical questions generated through our decomposition procedure, we augment the training data with ungrammatical samples.
Specifically, we add noise in the question by randomly dropping tokens with probability of $5\%$, and  replace~\whword~into `the' with probability of $5\%$.

\paragraph{Training Decomposition Scorer}
We create training data by making inferences for all reasoning types on \hotpot{} medium and hard examples.
We take the reasoning type that yields the correct answer as the gold reasoning type.
Appendix~\ref{app:impl-details} provides the full details.

\begin{table*}[!htb]
    \begin{minipage}{.5\linewidth}
      \centering
        \footnotesize
        \begin{tabularx}{0.95\textwidth}{X|c} 
         \toprule
         {Model} & F1 \\
         \midrule
            \sys & 70.57 $\rightarrow$ 59.07  \\
            \sys--1hop train & 61.73 $\rightarrow$ 58.30 \\
         \midrule
            \bert{} & 67.08 $\rightarrow$ 44.68  \\
            \bert{}--1hop train & 56.27  $\rightarrow$ 49.64  \\
         \bottomrule
        \end{tabularx}
    \end{minipage}
    \begin{minipage}{.5\linewidth}
      \centering
        \footnotesize
        \begin{tabularx}{0.95\textwidth}{X|c|c|c} 
         \toprule
          Model & Orig F1 & Inv F1 & Joint F1 \\
         \midrule
            \sys & 67.80 & 65.78 & 55.80 \\
            \bert & 54.65 & 32.49 & 19.27 \\
         \bottomrule
    \end{tabularx}
    \end{minipage} 
    \caption{
    \textbf{Left: modifying distractor paragraphs.} 
    F1 score on the original dev set and the new dev set made up with a different set of distractor paragraphs.
    \sys{} is our model and \sys{}--1hop train is \sys\ trained on only single-hop QA data and 400 decomposition annotations. \bert{} and \bert{}--1hop train are the baseline models, trained on \hotpot\ and single-hop data, respectively.
    \textbf{Right: adversarial comparison questions.} 
    F1 score on a subset of binary comparison questions. {\em Orig F1}, {\em Inv F1} and {\em Joint F1} indicate F1 score on the original example, the inverted example and the joint of two (example-wise minimum of two), respectively.
}\label{tab:robust}
\end{table*}

\subsection{Baseline Models}
We compare our system~\sys\ with the state-of-the-art on the \hotpot{} dataset as well as strong baselines. 

\vspace{.1cm}
\noindent \textbf{BiDAF} is \sota{} \RC{} model on \hotpot{}, originally from \citet{bidaf} and implemented by \citet{hotpot}.

\vspace{.1cm}
\noindent \textbf{\bert} is a large, language-model-pretrained model, achieving \sota{} results across many different NLP tasks~\citep{bert}.
This model is the same as our single-hop model described in Section~\ref{subsec:single}, but trained on the entirety of \hotpot{}.

\noindent \textbf{\bert--1hop train} is the same model but trained on single-hop QA data without \hotpot{} medium and hard examples.

\noindent \textbf{\sys--1hop train} is a variant of \sys\ that does not use multi-hop QA data except 400 decomposition annotations.
Since there is no access to the groundtruth answers of multi-hop questions, a decomposition scorer cannot be trained. Therefore, a final answer is obtained based on the confidence score from the single-hop \RC{} model, without a rescoring procedure.

\subsection{Results}

Table~\ref{tab:hotpot-result} compares the results of \sys\ with other baselines on the \hotpot{} development set.
We observe that \sys{} outperforms all baselines in both distractor and full wiki settings, outperforming the previous published result by a large margin. 
An interesting observation is that \sys{} not trained on multi-hop QA pairs (\sys--1hop train) shows reasonable performance across all data splits.

We also observe that \bert{} trained on single-hop \RC{} achieves a high F1 score, even though it does not draw inferences across different paragraphs.
For further analysis, we split the \hotpot{} development set into 
{single-hop solvable ({\em Single})} and {single-hop non-solvable ({\em Multi})}.\footnote{We consider an example to be solvable if all of three models of the \bert{}--1hop train ensemble obtains non-negative F1. This leads to 3426 single-hop solvable and 3979 single-hop non-solvable examples out of 7405 development examples, respectively.}
We observe that \sys{} outperforms \bert{} by a large margin in single-hop non-solvable ({\em Multi}) examples. This supports our attempt toward more explainable methods for answering multi-hop questions.

Finally, Table~\ref{tab:hotpot-test} shows the F1 score on the test set for distractor setting and full wiki setting on the leaderboard.\footnote{Retrieved on March 4th 2019 from \url{https://https://hotpotqa.github.io}}
These include unpublished models that are concurrent to our work. 
\sys\ achieves the best result out of models that report both distractor and full wiki setting.

\begin{table*}[tb]
    \centering
    \footnotesize
    \begin{tabular}{ll} 
     \toprule
        Question & Robert Smith founded the multinational company headquartered in what city? \\
     \midrule
        \multirow{2}{*}{Span-based} & Q1: Robert Smith founded which multinational company? \\
        & Q2: \answer~headquartered in what city? \\
     \midrule
        \multirow{2}{*}{Free-form} & Q1: Which multinational company was founded by Robert Smith? \\
        & Q2: Which city contains a headquarter of \answer?  \\
     \bottomrule
\end{tabular}
\caption{An example of the original question, span-based human-annotated \queries{} and free-form human-authored \queries.} 
\label{tab:example-queries}
\vspace{-8pt}
\end{table*}

\begin{table*}[!htb]\center
    \begin{minipage}{0.35\linewidth}
      \centering
      \footnotesize
      \begin{tabularx}{\textwidth}{Xc}
        \toprule
        \Queries{} & F1 \\
        \midrule
           Span (\pointer{c} trained on 200) & 65.44 \\
           Span (\pointer{c} trained on 400) & 69.44 \\
           Span (human) & 70.41 \\
           Free-form (human) & 70.76 \\
        \bottomrule
      \end{tabularx}
    \end{minipage}
    \hspace{2em}
    \begin{minipage}{0.4\linewidth}
      \centering
        \footnotesize
        \begin{tabularx}{\textwidth}{Xcc}
        \toprule
        Decomposition decision method & F1 \\
     \midrule
        Confidence-based & 61.73 \\
        Pipeline & 63.59 \\
        Decomposition scorer (\sys) & 70.57 \\
        Oracle & 76.75 \\
     \bottomrule
        \end{tabularx}
    \end{minipage} 
    \caption{
    \textbf{Left: ablations in \queries.} 
    F1 score on a sample of 50 bridging questions from the dev set of \hotpot{}, \pointer{c} is our span-based model trained with 200 or 400 annotations.
    \textbf{Right: ablations in decomposition decision method.} 
    F1 score on the dev set of \hotpot{}  with ablating decomposition decision method. Oracle indicates that the ground truth reasoning type is selected.
}\label{tab:ablation}
\end{table*}

\subsection{Evaluating Robustness}\label{subsec:robustness}
In order to evaluate the robustness of different methods to changes in the data distribution, we set up two adversarial settings in which the trained model remains the same but the evaluation dataset is different.

\paragraph{Modifying Distractor Paragraphs.} We collect a new set of distractor paragraphs to evaluate if the models are robust to the change in distractors.\footnote{We choose 8 distractor paragraphs that do not to change the groundtruth answer.} In particular, we follow the same strategy as the original approach~\citep{hotpot} using TF-IDF similarity between the question and the paragraph, but with no overlapping distractor paragraph with the original distractor paragraphs.
Table~\ref{tab:robust} compares the F1 score of \sys\ and \bert{} in the original distractor setting and in the modified distractor setting.
As expected, the performance of both methods degrade, but \sys\ is more robust to the change in distractors.
Namely, \sys{}--1hop train degrades much less (only 3.41 F1) compared to other approaches because it is only trained on single-hop data and therefore does not exploit the data distribution. 
These results confirm our hypothesis that the end-to-end model is sensitive to the change of the data and our model is more robust.


\paragraph{Adversarial Comparison Questions.}
We create an adversarial set of comparison questions by altering the original question  so that the correct answer is inverted.
For example, we change \example{Who was born earlier, Emma Bull or Virginia Woolf?} to \example{Who was born later, Emma Bull or Virginia Woolf?}
We automatically invert 665 questions (details in Appendix~\ref{app:invcomp}). 
We report the joint F1, taken as the minimum of the prediction F1 on the original and the inverted examples.
Table~\ref{tab:robust} shows the joint F1 score of \sys\ and \bert{}.
We find that \sys\ is robust to inverted questions, and outperforms \bert{} by 36.53 F1.

\subsection{Ablations}\label{subsec:ablation}
\paragraph{Span-based vs. Free-form \queries{}.}
We evaluate the quality of generated \queries{} using span-based question decomposition. We replace the question decomposition component using \pointer{3} with (i) \query{} decomposition through groundtruth spans, (ii) \query{} decomposition with free-form, hand-written \queries{} (examples shown in Table~\ref{tab:example-queries}).

Table~\ref{tab:ablation} (left) compares the question answering performance of \sys\ when replaced with alternative \queries{} on a sample of 50 bridging questions.\footnote{A full set of samples is shown in Appendix~\ref{app:analysis}.}
There is little difference in model performance between span-based and \queries{} written by human.
This indicates that our span-based \queries{} are as effective as free-form \queries{}.
In addition, \pointer{3}~trained on 200 or 400 examples obtains close to human performance.
We think that identifying spans often rely on syntactic information of the question, which \bert{} has likely learned from language modeling.
We use the model trained on 200 examples for \sys{} to demonstrate sample-efficiency, and expect performance improvement with more annotations.


\vspace{-.1cm}
\paragraph{Ablations in decomposition decision method.}
Table~\ref{tab:ablation} (right) compares different ablations to evaluate the effect of the decomposition scorer.
For comparison, we report the F1 score of the confidence-based method which chooses the decomposition with the maximum confidence score from the single-hop RC model, and the pipeline approach which independently selects the reasoning type as described in Section~\ref{subsec:recombination}.
In addition, we report an oracle which takes the maximum F1 score across different reasoning types to provide an upperbound. 
A pipeline method gets lower F1 score than the decomposition scorer.
This suggests that using more context from decomposition (e.g., the answer and the evidence) helps avoid cascading errors from the pipeline.
Moreover, a gap between \sys\ and oracle (6.2 F1) indicates that there is still room to improve.

\begin{table}[t]
    \centering
    \small
    \begin{tabular}{lc}
        \toprule
        \multicolumn{2}{c}{Breakdown of 15 failure cases} \\
        \midrule
        Incorrect groundtruth & 1  \\
        Partial match with the groundtruth & 3 \\
        Mistake from human & 3 \\
        Confusing question & 1\\
        \Query{} requires cross-paragraph reasoning & 2\\
        Decomposed \queries{} miss some information & 2  \\
        Answer to the first \query{} can be multiple & 3\\
        \bottomrule              
    \end{tabular}
    \caption{
    The error analyses of human experiment, where the upperbound F1 score of span-based \queries{} with no decomposition scorer is measured.
    }
    \label{tab:upperbound-breakdown}
\end{table}

\begin{table*}[t]
    \centering
    \small
    \begin{tabularx}{\textwidth}{X}
        \toprule
        \textbf{Q} What country is the Selun located in? \\
        \textbf{P1} Selun lies between the valley of Toggenburg and Lake Walenstadt in the canton of St. Gallen. \\
        \textbf{P2} The canton of St. Gallen is a canton of {\color{red}Switzerland}. \\
        \midrule
        \textbf{Q} Which pizza chain has locations in more cities, Round Table Pizza or Marion's Piazza? \\
        \textbf{P1} {\color{red}Round Table Pizza} is a large chain of pizza parlors in the western United States. \\
        \textbf{P2} Marion's Piazza ... the company currently operates 9 restaurants throughout the greater Dayton area. \\
        \textbf{Q1} Round Table Pizza has locations in how many cities?
        \textbf{Q2} Marion 's Piazza has locations in how many cities? \\
        \midrule
        \textbf{Q} Which magazine had more previous names, Watercolor Artist or The General?\\
        \textbf{P1} Watercolor Artist, formerly Watercolor Magic, is an American bi-monthly magazine that focuses on ...  \\
        \textbf{P2} {\color{red}The General} (magazine): Over the years the magazine was variously called `The Avalon Hill General', `Avalon Hill's General', `The General Magazine', or simply `General'. \\
        \textbf{Q1} Watercolor Artist had how many previous names? \textbf{Q2} The General had how many previous names? \\
        \bottomrule              
    \end{tabularx}
    \caption{
    The failure cases of \sys, where {Q}, {P1} and {P2} indicate the given question and paragraphs, and {Q1} and {Q2} indicate \queries{} from \sys.
    (Top) The required multi-hop reasoning is implicit, and the question cannot be decomposed. (Middle) \sys{} decomposes the question well but fails to answer the first \query{} because there is no explicit answer. (Bottom) \sys{} is incapable of counting.
    }
    \label{tab:error}
\end{table*}

\paragraph{Upperbound of Span-based \Queries{} without a decomposition scorer.}
To measure an upperbound of span-based \queries{} without a decomposition scorer, where a human-level RC model is assumed, we conduct a human experiment on a sample of 50 bridging questions.\footnote{A full set of samples is shown in Appendix~\ref{app:analysis}.} In this experiment, humans are given each \query{} from decomposition annotations and are asked to answer it without an access to the original, multi-hop question.
They are asked to answer each \query{} with no cross-paragraph reasoning, and mark it as a failure case if it is impossible.
The resulting F1 score, calculated by replacing RC model to humans, is 72.67 F1.

Table~\ref{tab:upperbound-breakdown} reports the breakdown of fifteen error cases. 53\% of such cases are due to the incorrect groundtruth, partial match with the groundtruth or mistake from humans. 47\% are genuine failures in the decomposition. For example, a multi-hop question ``Which animal races annually for a national title as part of a post-season NCAA  Division I Football Bowl Subdivision college football game?" corresponds to the last category in Table~\ref{tab:upperbound-breakdown}. The question can be decomposed into ``Which post-season NCAA Division I Football Bowl Subdivision college football game?" and
``Which animal races annually for a national title as part of \answer{}?". However in the given set of paragraphs, there are multiple games that can be the answer to the first \query{}. Although only one of them is held with the animal racing, it is impossible to get the correct answer only given the first \query{}. We think that incorporating the original question along with the \queries{} can be one solution to address this problem, which is partially done by a decomposition scorer in \sys.

\vspace{-.1cm}
\paragraph{Limitations.}
We show the overall limitations of \sys\ in Table~\ref{tab:error}. First, some questions are not compositional but require implicit multi-hop reasoning, hence cannot be decomposed. Second, there are questions that can be decomposed but the answer for each \query{} does not exist explicitly in the text, and must instead by inferred with commonsense reasoning. Lastly, the required reasoning is sometimes beyond our reasoning types (e.g. counting or calculation). Addressing these remaining problems is a promising area for future work.
\section{Conclusion}\label{sec:concl}
We proposed \sys, a system for multi-hop \RC{} that decomposes a multi-hop question into simpler, single-hop \queries{}.
We recasted \query{} generation as a span prediction problem, allowing the model to be trained on 400 labeled examples to generate high quality \queries{}.
Moreover, \sys{} achieved further gains from the decomposition scoring step.
\sys{} achieved \sota{} on \hotpot{} distractor setting and full wiki setting, while providing explainable evidence for its decision making in the form of \queries{} and being more robust to adversarial settings than strong baselines.

\section*{Acknowledgments}
This research was supported by ONR (N00014-18-1-2826, N00014-17-S-B001), NSF (IIS 1616112, IIS 1252835, IIS 1562364), ARO (W911NF-16-1-0121), an Allen Distinguished Investigator Award, Samsung GRO and gifts from Allen Institute for AI, Google, and Amazon. 

We thank the anonymous reviewers and UW NLP members for their thoughtful comments and discussions.

\bibliography{journal-abbrv,bib}
\bibliographystyle{acl_natbib}

\clearpage
\appendix
\section{Span Annotation}\label{app:annotation}\begin{figure}[!thb]
\centering
\resizebox{\columnwidth}{!}{\includegraphics[width=\textwidth,trim={0 0 4cm 0},clip]{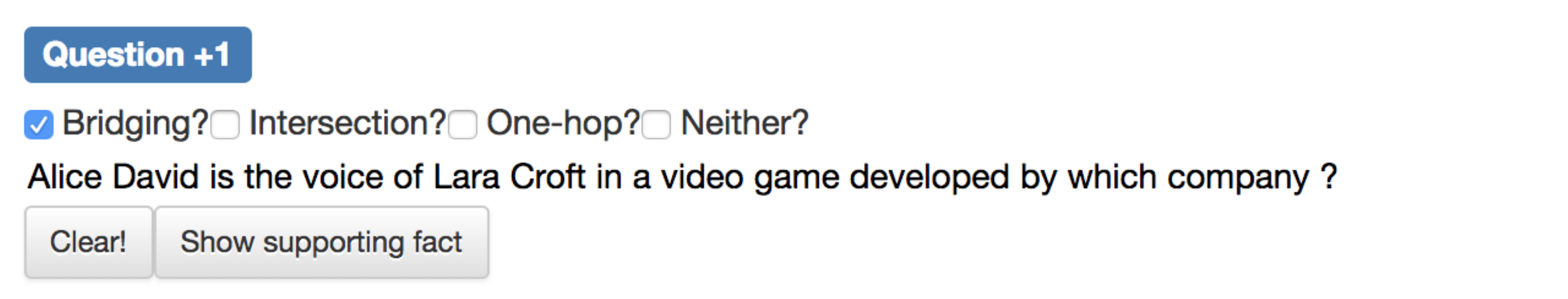}}
\resizebox{\columnwidth}{!}{\includegraphics[width=\textwidth,trim={0 0 4cm 0},clip]{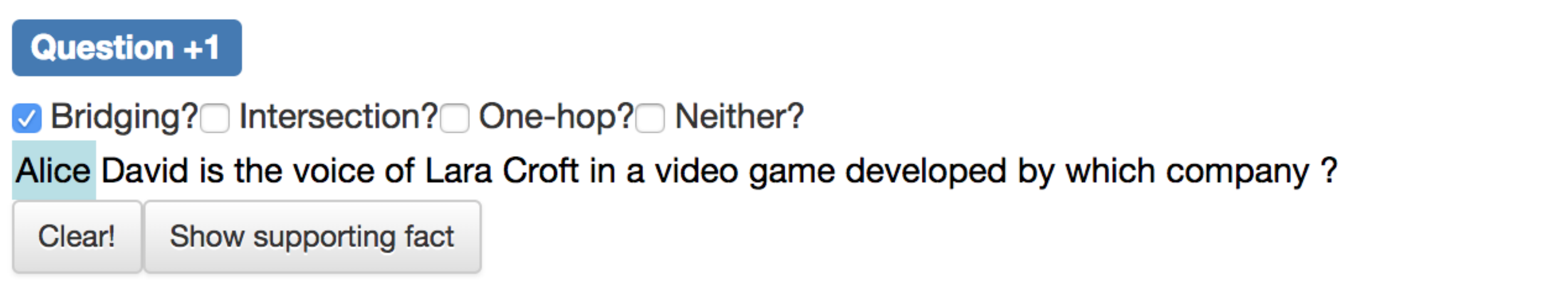}}
\resizebox{\columnwidth}{!}{\includegraphics[width=\textwidth,trim={0 0 4cm 0},clip]{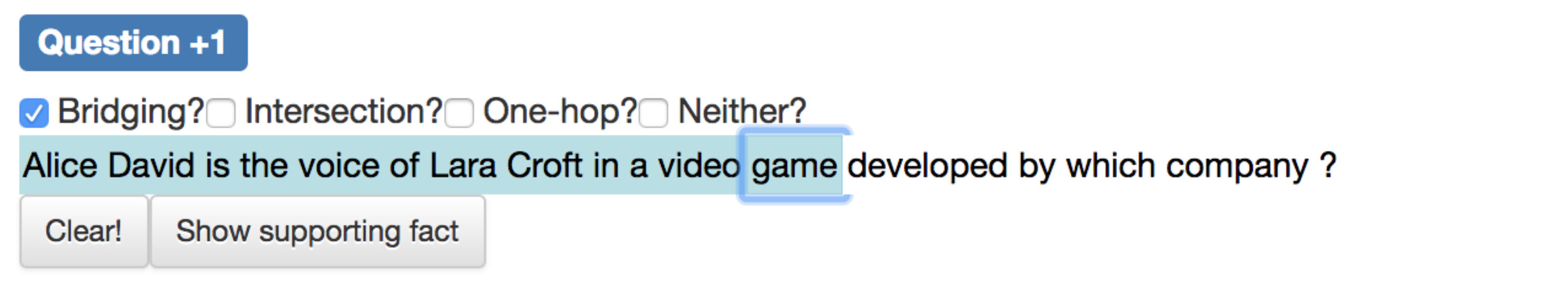}}
\resizebox{\columnwidth}{!}{\includegraphics[width=\textwidth,trim={0 0 4cm 0},clip]{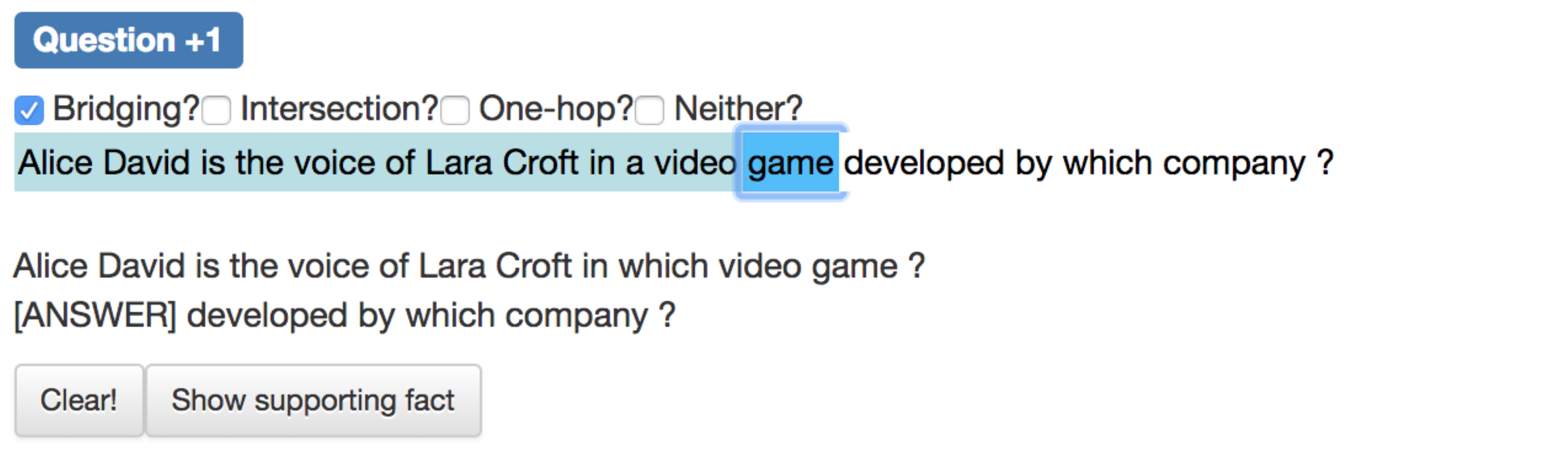}}
\resizebox{\columnwidth}{!}{\includegraphics[width=\textwidth,trim={0 0 4cm 0},clip]{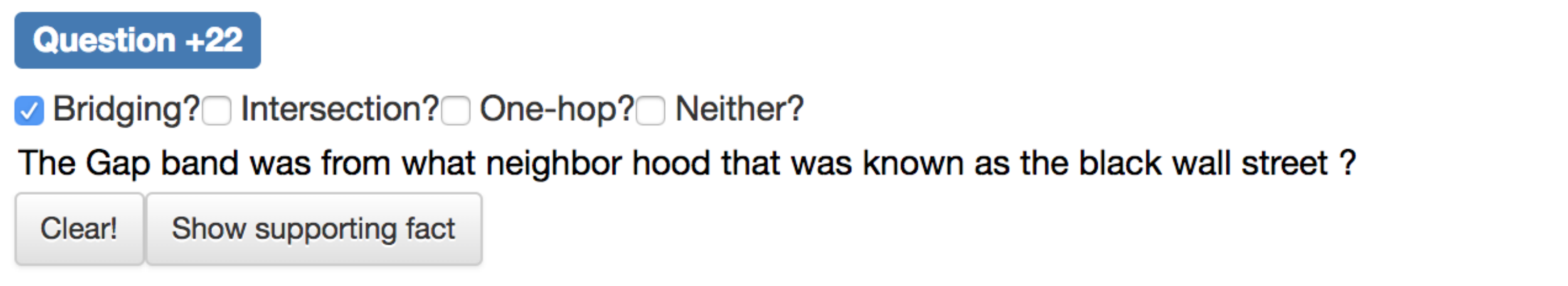}}
\resizebox{\columnwidth}{!}{\includegraphics[width=\textwidth,trim={0 0 4cm 0},clip]{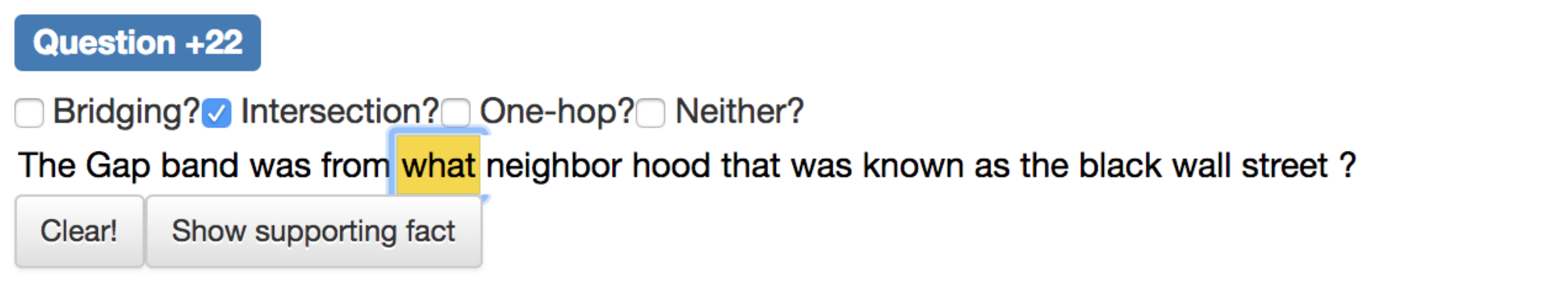}}
\resizebox{\columnwidth}{!}{\includegraphics[width=\textwidth,trim={0 0 4cm 0},clip]{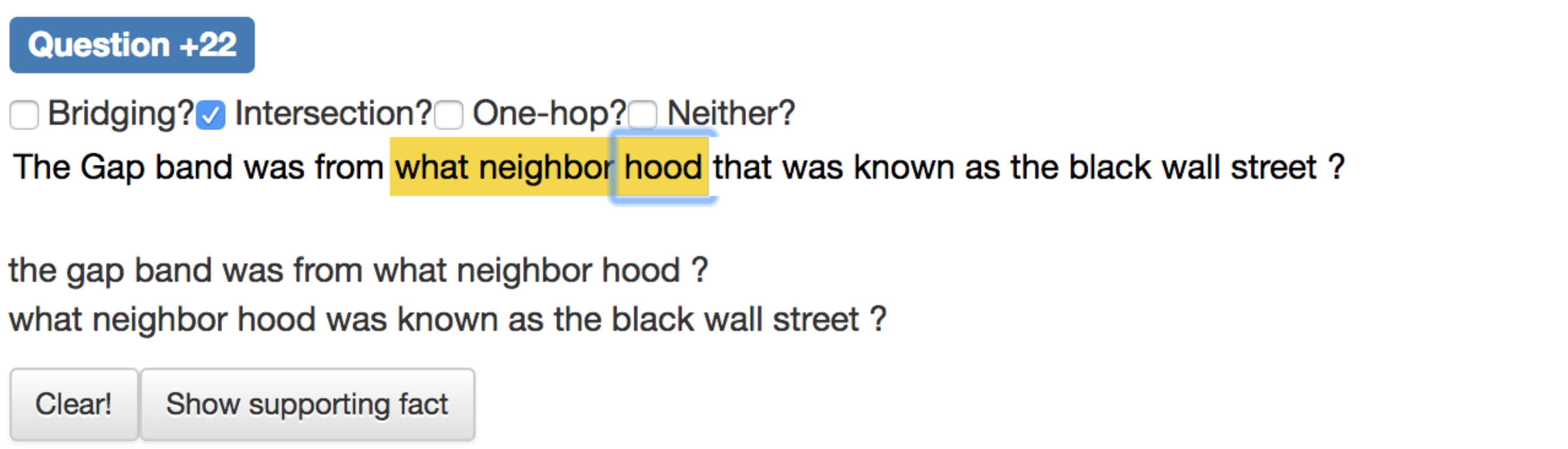}}
\caption{
Annotation procedure. Top four figures show annotation for bridging question. Bottom three figures show annotation for intersection question. 
}
\label{fig:annotation-procedure}
\end{figure}

In this section, we describe span annotation collection procedure for bridging and intersection questions.

The goal is to collect three points (bridging) or two points (intersection) given a multi-hop question.
We design an interface to annotate span over the question by clicking the word in the question. First, given a question, the annotator is asked to identify which reasoning type out of bridging, intersection, one-hop and neither is the most proper.\footnote{Note that we exclude comparison questions for annotations, since comparison questions are already labeled on \hotpot.} Since bridging type is the most common, bridging is checked by default.
If the question type is bridging, the annotator is asked to make three clicks for the start of the span, the end of the span, and the head-word (top four examples in Figure~\ref{fig:annotation-procedure}). After three clicks are all made, the annotator can see the heuristically generated \queries{}.
If the question type is intersection, the annotator is asked to make two clicks for the start and the end of the second segment out of three segments (bottom three examples in Figure~\ref{fig:annotation-procedure}). Similarly, the annotator can see the heuristically generated \queries{} after two clicks.
If the question type is one-hop or neither, the annotator does not have to make any click.
If the question can be decomposed into more than one way, the annotator is asked to choose the more natural decomposition.
If the question is ambiguous, the annotator is asked to pass the example, and only annotate for the clear cases.
For the quality control, all annotators have enough in person, one-on-one tutorial sessions and are given 100 example annotations for the reference.

\section{Decompotision for Comparison}\label{app:heuristics}\begin{table*}[t]
    \centering
    \footnotesize
    \begin{tabular}{l} 
     \toprule
     Operation \& Example \\
     \midrule
        \textbf{Type: Numeric}\\
        Is greater (\answer) (\answer) $\rightarrow$ \yes~or~\no\\
        Is smaller (\answer) (\answer) $\rightarrow$ \yes~or~\no \\
        Which is greater (\keyword,~\answer) (\keyword,~\answer) $\rightarrow$ \keyword\\
        Which is smaller (\keyword,~\answer) (\keyword,~\answer) $\rightarrow$ \keyword \\
        \\
        Did \hl{the Battle of Stones River} occur before \hl{the Battle of Saipan}? \\
        \ \ \ Q1: The Battle of Stones River occur when? $\rightarrow$ 1862 \\
        \ \ \ Q2: The Battle of Saipan River occur when? $\rightarrow$ 1944 \\
        \ \ \ Q3: Is smaller (the Battle of Stones River, 1862) (the Battle of Saipan, 1944) $\rightarrow$ \yes \\
     \midrule
        \textbf{Type: Logical}\\
        And (\answer) (\answer)  $\rightarrow$~\yes~or~\no\\
        Or (\answer) (\answer) $\rightarrow$~\yes~or~\no\\
        Which is true (\keyword,~\answer) (\keyword,~\answer)  $\rightarrow$ \keyword \\
        \\
        In between \hl{Atsushi Ogata} and \hl{Ralpha Smart} who graduated from Harvard College?\\
        \ \ \ Q1: Atsushi Ogata graduated from Harvard College? $\rightarrow$ \yes \\
        \ \ \ Q2: Ralpha Smart graduated from Harvard College? $\rightarrow$ \no \\
        \ \ \ Q3: Which is true (Atsushi Ogata,~\yes) (Ralpha Smart,~\no) $\rightarrow$ Atsushi Ogata \\
     \midrule
        \textbf{Type: String}\\
        Is equal (\answer) (\answer)  $\rightarrow$ \yes~or~\no\\
        Not equal (\answer) (\answer) $\rightarrow$ \yes~or~\no\\
        Intersection (\answer) (\answer) $\rightarrow$ string \\
        \\
        Are \hl{Cardinal Health} and \hl{Kansas City Southern} located in the same state? \\
        \ \ \ Q1: Cardinal Health located in which state? $\rightarrow$ Ohio \\
        \ \ \ Q2: Cardinal Health located in which state? $\rightarrow$ Missouri \\
        \ \ \ Q3: Is equal (Ohio) (Missouri) $\rightarrow$~\no \\
     \bottomrule
\end{tabular}
\caption{A set of discrete operations proposed for comparison questions, along with the example on each type. \answer~is the answer of each query, and~\keyword~is the entity corresponding to each query. The answer of each query is shown in the right side of $\rightarrow$. If the question and two entities for comparison are given, queries and a discrete operation can be obtained by heuristics.} 
\label{tab:comparison}
\vspace{-8pt}
\end{table*}

In this section, we describe the decomposition procedure for comparison, which does not require any extra annotation.

Comparison requires to compare a property of two different entities, usually requiring discrete operations.
We identify 10 discrete operations which sufficently cover comparison operations, shown in Table~\ref{tab:comparison}.
Based on these pre-defined discrete operations, we decompose the question through the following three steps.

First, we extract two entities under comparison. We use \pointer{4} to obtain $\mathrm{ind}_1,  \dots, \mathrm{ind}_4$, where $\mathrm{ind}_1$ and $\mathrm{ind}_2$ indicate the start and the end of the first entity, and $\mathrm{ind}_3$ and $\mathrm{ind}_4$ indicate those of the second entity. We create a training data which each example contains the question and four points as follows: we filter out bridge questions in \hotpot{} to leave comparison questions, extract the entities using Spacy\footnote{\url{https://spacy.io/}} NER tagger in the question and in two supporting facts (annotated sentences in the dataset which serve as evidence to answer the question), and match them to find two entities which appear in one supporting sentence but not in the other supporting sentence. 

Then, we identity the suitable discrete operation, following Algorithm~\ref{alg:comparison-heuristics}.

Finally, we generate \queries{} according to the discrete operation. Two \queries{} are obtained for each entity.

\begin{algorithm*}
\small
\caption{Algorithm for Identifying Discrete Operation.
First, given two entities for comparison, the coordination and the preconjunct or the predeterminer are identified.
Then, the quantitative indicator and the head entity is identified if they exist, where a set of uantitative indicators is pre-defined. In case any quantitative indicator exists, the discrete operation is determined as one of numeric operations. If there is no quantitative indicator, the discrete operation is determined as one of logical operations or string operations. 
}\label{alg:comparison-heuristics}
\begin{algorithmic}
\Procedure{Find\_Operation}{question, entity1, entity2}
\State coordination, preconjunct $\gets f$(question, entity1, entity2) 
\State Determine if the question is {\em either} question or {\em both} question from coordination and preconjunct
\State head entity $\gets f_{head}$(question, entity1, entity2)
\If {{\em more, most, later, last, latest, longer, larger, younger, newer, taller, higher} in question}
    \If {head entity exists}
        discrete\_operation $\gets$ Which is greater
    \Else{}
        discrete\_operation $\gets$ Is greater
    \EndIf {}
\ElsIf {{\em less, earlier, earliest, first, shorter, smaller, older, closer} in question}   
    \If {head entity exists}
        discrete\_operation $\gets$ Which is smaller
    \Else{}
        discrete\_operation $\gets$ Is smaller
    \EndIf {}
\ElsIf {head entity exists}
    \State discrete\_operation $\gets$ Which is true
\ElsIf {question is not yes/no question and asks for the property in common}
    \State discrete\_operation $\gets$ Intersection
\ElsIf {question is yes/no question}
    \State Determine if question asks for logical comparison or string comparison
    \If {question asks for logical comparison}
        \If {{\em either} question}
        discrete\_operation $\gets$ Or
        \ElsIf {{\em both} question}
            discrete\_operation $\gets$ And
        \EndIf {}
    \ElsIf {question asks for string comparison}
        \If {asks for same?}
            discrete\_operation $\gets$ Is equal
        \ElsIf {asks for difference?}
            discrete\_operation $\gets$ Not equal
        \EndIf {}
    \EndIf {}
\EndIf
\State \Return discrete\_operation
\EndProcedure
\end{algorithmic}
\end{algorithm*}

\section{Implementation Details}\label{app:impl-details}\paragraph{Implementation Details.}
We use PyTorch \citep{pytorch} on top of Hugging Face's~\bert~implementation.\footnote{\url{https://github.com/huggingface/pytorch-pretrained-BERT}} We tune our model from Google's pretrained \bertbase~(lowercased)\footnote{\url{https://github.com/google-research/bert}}, containing 12 layers of Transformers~\citep{transformer} and a hidden dimension of 768.
We optimize the objective function using Adam~\citep{adam} with learning rate $5\times10^{-5}$.
We lowercase the input and set the maximum sequence length $\vert{S}\vert$ to $300$ for models which input is both the question and the paragraph, and $50$ for the models which input is the question only.
\section{Creating Inverted Binary Comparison Questions}\label{app:invcomp}We identify the comparison question with 7 out of 10 discrete operations (Is greater, Is smaller, Which is greater, Which is smaller, Which is true, Is equal, Not equal) can automatically be inverted. It leads to 665 inverted questions.

\section{A Set of Samples used for Ablations}\label{app:analysis}\begin{table*}[t]
    \centering
    \scriptsize
    \begin{tabular}{c}
        \toprule
        5abce73055429959677d6b34,5a80071f5542992bc0c4a684,5a840a9e5542992ef85e2397,5a7e02cf5542997cc2c474f4,5ac1c9a15542994ab5c67e1c\\
        5a81ea115542995ce29dcc78,5ae7308d5542991e8301cbb8,5ae527945542993aec5ec167,5ae748d1554299572ea547b0,5a71148b5542994082a3e567\\
        5ae531695542990ba0bbb1fb,5a8f5273554299458435d5b1,5ac2db67554299657fa290a6,5ae0c7e755429945ae95944c,5a7150c75542994082a3e7be\\
        5abffc0d5542990832d3a1e2,5a721bbc55429971e9dc9279,5ab57fc4554299488d4d99c0,5abbda84554299642a094b5b,5ae7936d5542997ec27276a7\\
        5ab2d3df554299194fa9352c,5ac279345542990b17b153b0,5ab8179f5542990e739ec817,5ae20cd25542997283cd2376,5ae67def5542991bbc9760f3\\
        5a901b985542995651fb50b0,5a808cbd5542996402f6a54b,5a84574455429933447460e6,5ab9b1fd5542996be202058e,5a7f1ad155429934daa2fce2\\
        5ade03da5542997dc7907120,5a809fe75542996402f6a5ba,5ae28058554299495565da90,5abd09585542996e802b469b,5a7f9cbd5542994857a7677c\\
        5a7b4073554299042af8f733,5ac119335542992a796dede4,5a7e1a2955429965cec5ea5d,5a8febb555429916514e73e4,5a87184a5542991e771816c5\\
        5a86681c5542991e77181644,5abba584554299642a094afa,5add39e75542997545bbbcc4,5a7f354b5542992e7d278c8c,5a89810655429946c8d6e929\\
        5a78c7db55429974737f7882,5a8d0c1b5542994ba4e3dbb3,5a87e5345542993e715abffb,5ae736cb5542991bbc9761c2,5ae057fd55429945ae959328\\
        \bottomrule              
    \end{tabular}
    \caption{
    Question IDs from a set of samples used for ablations in Section~\ref{subsec:ablation}.
    }
    \label{tab:samples}
\end{table*}

A set of samples used for ablations in Section~\ref{subsec:ablation} is shown in Table~\ref{tab:samples}.

\end{document}